\documentclass[letterpaper]{article}
\usepackage{aaai2026} 
\usepackage{times}
\usepackage{helvet}
\usepackage{courier}
\usepackage[hyphens]{url}
\usepackage{graphicx}
\usepackage{natbib}
\usepackage{amsmath}
\usepackage{caption}
\usepackage{booktabs}
\usepackage[table]{xcolor}
\usepackage{multirow}
\usepackage{float}
\usepackage{pifont}
\makeatletter
\@ifundefined{nocopyright}{}{%
  \nocopyright
}
\providecommand{\@copyrighttext}{}

\providecommand{\@copyrightspace}{}
\makeatother
\title{Forecast then Calibrate: Feature Caching as ODE for\\Efficient Diffusion Transformers}

\author{
    Shikang Zheng\textsuperscript{\rm 1,2}, 
    Liang Feng\textsuperscript{\rm 1,3}, 
    Xinyu Wang\textsuperscript{\rm 1}, 
    Qinming Zhou\textsuperscript{\rm 1,4}, 
    Peiliang Cai\textsuperscript{\rm 1}, 
    Chang Zou\textsuperscript{\rm 1}, 
    Jiacheng Liu\textsuperscript{\rm 1}, 
    Yuqi Lin\textsuperscript{\rm 1}, 
    Junjie Chen\textsuperscript{\rm 1}, 
    Yue Ma\textsuperscript{\rm 5}, 
    Linfeng Zhang\textsuperscript{\rm 1,\dag} \\
    \normalfont
    \textsuperscript{\rm 1} Shanghai Jiao Tong University \quad
    \textsuperscript{\rm 2} South China University of Technology \quad
    \textsuperscript{\rm 3} Fudan University \\
    \textsuperscript{\rm 4} Tsinghua University \quad
    \textsuperscript{\rm 5} Hong Kong University of Science and Technology
}

\begin{document}

\maketitle

% 首页脚注：† denotes the corresponding author.
\renewcommand{\thefootnote}{\fnsymbol{footnote}}
\footnotetext{${}^\dag$ Denotes the corresponding author.}
\renewcommand{\thefootnote}{\arabic{footnote}}

% Include sections from separate files
\begin{abstract}
Diffusion Transformers (DiTs) have demonstrated exceptional performance in high-fidelity image and video generation. To reduce their substantial computational costs, feature caching techniques have been proposed to accelerate inference by reusing hidden representations from previous timesteps. However, current methods often struggle to maintain generation quality at high acceleration ratios, where prediction errors increase sharply due to the inherent instability of long-step forecasting. In this work, we adopt an ordinary differential equation (ODE) perspective on the hidden-feature sequence, modeling layer representations along the trajectory as a feature-ODE. We attribute the degradation of existing caching strategies to their inability to robustly integrate historical features under large skipping intervals. To address this, we propose \textbf{FoCa} (Forecast-then-Calibrate), which treats feature caching as a feature-ODE solving problem. Extensive experiments on image synthesis, video generation, and super-resolution tasks demonstrate the effectiveness of FoCa, especially under aggressive acceleration. Without additional training, FoCa achieves near-lossless speedups of 5.50$\times$ on FLUX, 6.45$\times$ on HunyuanVideo, 3.17$\times$ on Inf-DiT, and maintains high quality with a 4.53$\times$ speedup on DiT. \emph{Our code will be released upon acceptance.}
\end{abstract}
\section{Introduction}\label{sec:introduction}

Recent progress in diffusion-based generative models has substantially advanced the capabilities of visual synthesis, achieving unprecedented quality in both image~\citep{DDPM} and video generation~\citep{blattmann2023stablevideodiffusionscaling}. In particular, Diffusion Transformers (DiTs)~\citep{DiT} have set new performance standards by leveraging transformer-based architectures, which offer superior scalability and modeling capacity. Despite these advantages, DiTs remain computationally intensive due to their iterative sampling process, where each output requires a sequence of denoising steps that scale linearly with both the number of steps and model complexity. This computational overhead presents significant barriers to real-time or resource-constrained deployment.

\begin{figure}[htbp]
  \centering
  \includegraphics[trim=25 25 25 20, clip, width=1\linewidth]{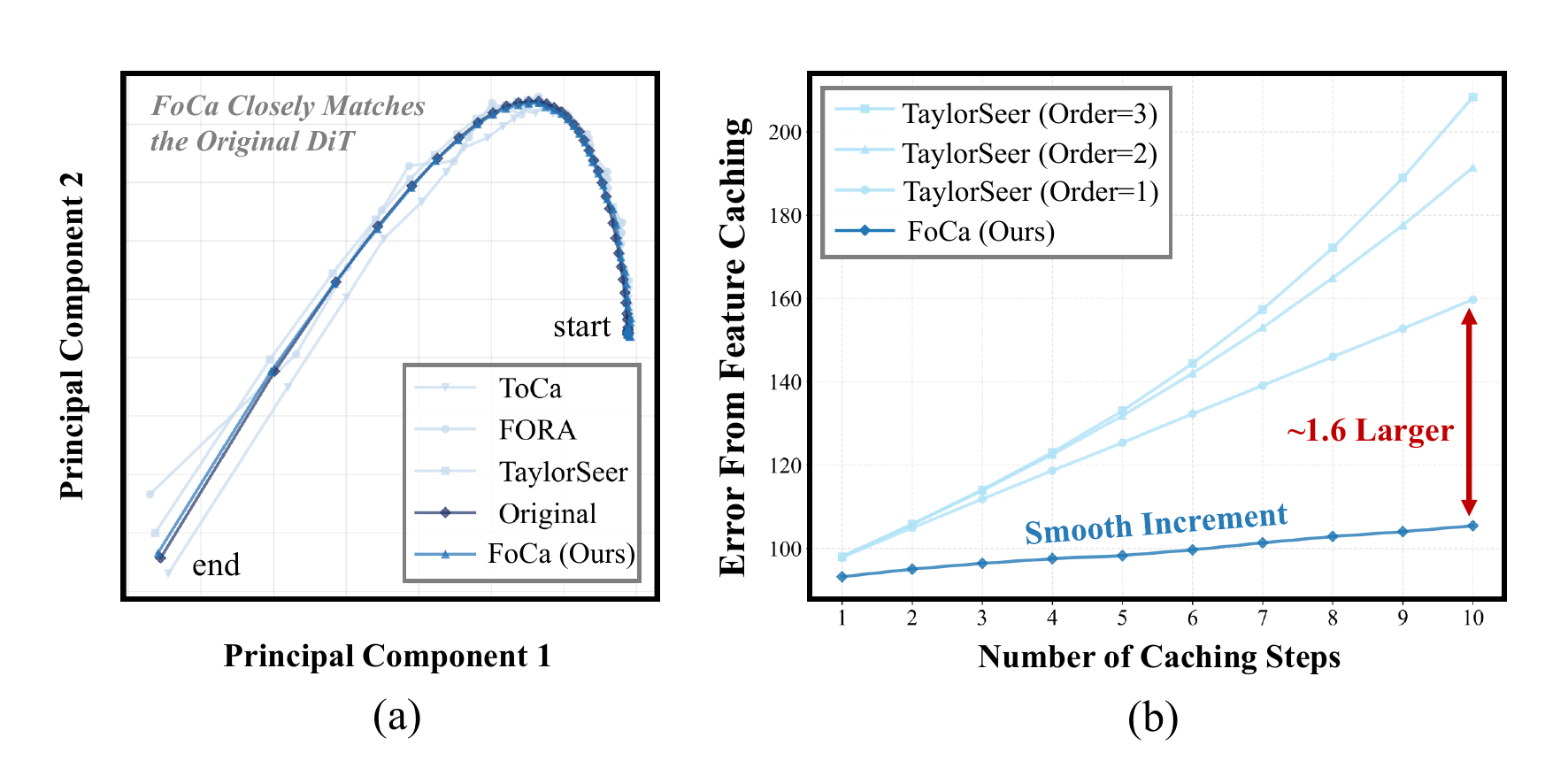}
  \caption{(a) PCA visualization of feature trajectory: FoCa closely follows the original model’s path. (b) Prediction error compared with the original model: FoCa maintains significantly lower error, whereas TaylorSeer’s higher‐order forecasts incur rapidly growing errors.}
  \label{fig:vs1}
\end{figure}

To alleviate this bottleneck, two main acceleration strategies have been proposed: reducing the number of sampling steps through algorithmic enhancements~\citep{ma2023deepcacheacceleratingdiffusionmodels}, and minimizing per-step computational cost via architectural modifications~\citep{yuan2024ditfastattnattentioncompressiondiffusion, zhao2025realtimevideogenerationpyramid}. Among these, training-free \emph{feature caching} has emerged as a promising and general-purpose solution. It builds on the observation that intermediate hidden representations exhibit temporal consistency. This temporal coherence allows the model to bypass redundant computations by reusing or predicting features at selected timesteps, effectively reducing latency without retraining the network.

Initial implementations of feature caching were developed in the context of U-Net architectures, where cached activations are reused across nearby timesteps using skip connections~\citep{wimbauer2024cachecanacceleratingdiffusion, ma2023deepcacheacceleratingdiffusionmodels}. As DiTs become the dominant architecture for high-fidelity generation, recent works have extended caching mechanisms to this setting, leading to methods such as FORA, ToCa, and TaylorSeer~\citep{FORA, ToCa, TaylorSeer}. These techniques leverage temporal token similarity to inform feature reuse or extrapolation, but still suffer from error accumulation and instability under long skip intervals.

\begin{figure*}[t]
  \centering
  \includegraphics[trim=10 20 10 15, clip,width=\textwidth]{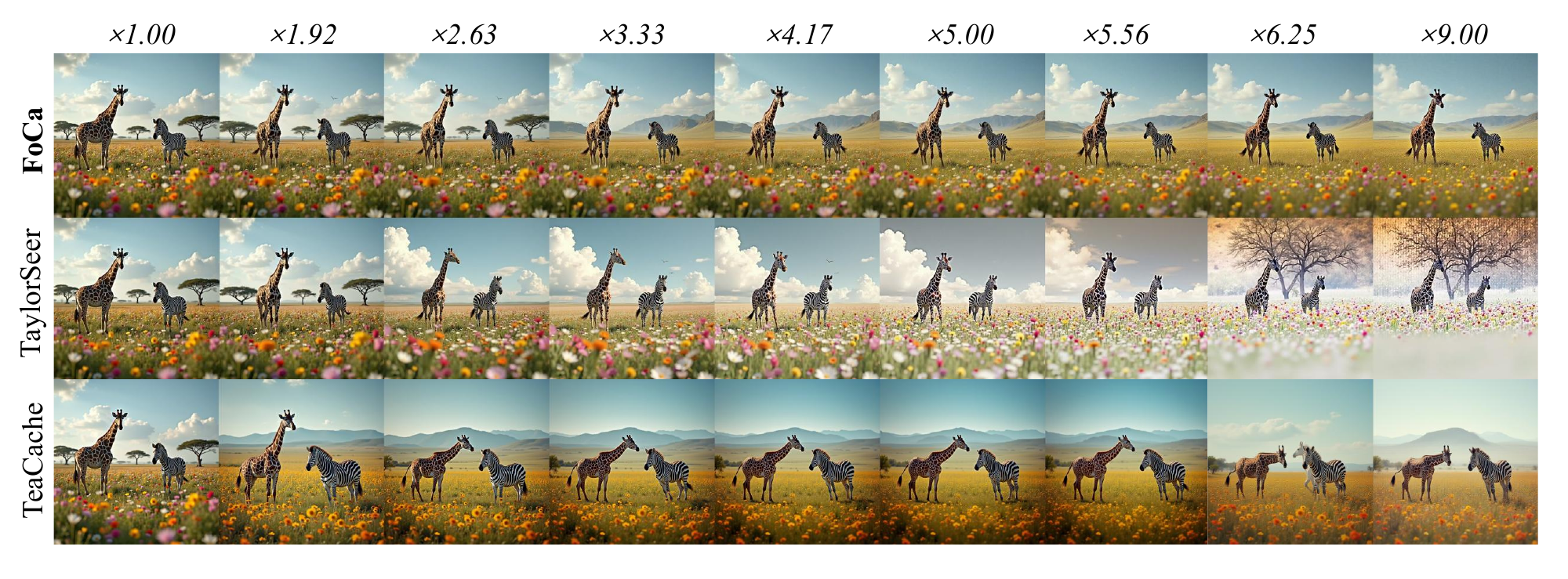}
  \caption{Visualization of the image generated by different methods on prompt: \emph{``Zoomed out view of a giraffe and a zebra in the middle of a field covered with colorful flowers.''} As the computation ratio increases, the image quality of TaylorSeer and TeaCache degrades, whereas FoCa maintains better performance.}
  \label{fig:sequence}
\end{figure*}

\textbf{Limited Utilization of Historical Information.}
Existing methods fail to fully exploit historical hidden states for stable prediction. Reuse-only approaches simply replicate the most recent feature without incorporating multi-step history or adapting to the evolving feature dynamics over time—particularly the shifting eigenvalue spectrum of the linearized drift. Forecasting-based methods such as TaylorSeer attempt to transform historical information into local approximations by estimating time derivatives via finite differences and applying truncated Taylor expansions. However, this strategy introduces significant numerical errors, as the finite-difference estimation is highly sensitive to noise and becomes unstable in higher-order terms. Consequently, neither reuse nor current forecasting methods adequately leverage historical information for robust and accurate prediction under large skipping intervals.

\textbf{Instability Under Large Skipping Intervals.}
As the interval between executed timesteps increases, existing caching methods struggle to maintain accurate hidden state estimation. In reuse-based approaches, simply copying outdated features leads to growing mismatches as temporal distance increases. Similarly, forecasting methods such as Taylor series extrapolation suffer from accumulated approximation errors when extrapolating over longer spans. These inaccuracies compound with larger skip lengths, resulting in significant deviations from the correct denoising trajectory. Empirically, this leads to rapid degradation in generation quality, as shown in Figure~\ref{fig:sequence}, limiting the practical effectiveness of current caching strategies.

To solve these problems, we reformulate feature caching as solving a feature-ODE, providing a principled approach to modeling hidden-state evolution. This perspective enables the application of multi-step solvers that leverage historical hidden states directly, without relying on high-order derivative approximations. Specifically, we introduce \textbf{FoCa} (Forecast-then-Calibrate) framework, which combines a Backward Differentiation Formula based predictor with a lightweight Heun corrector. The forecast component aggregates previous features for stable forecasting, while the Heun calibration incorporates the previous full activation as a reference, damping overshoot on larger skip intervals.

FoCa provides robust and efficient prediction across diverse tasks and architectures. Without additional training or architectural changes, it achieves near-lossless acceleration of \textbf{5.50$\times$} on FLUX for text-to-image generation, \textbf{6.45$\times$} on HunyuanVideo for text-to-video generation, and \textbf{3.17$\times$} on Inf-DiT for super-resolution. FoCa also maintains high image quality with \textbf{4.53$\times$} acceleration on DiT-XL/2. In summary, our main contributions are:

\begin{figure}[t]
  \centering
  \includegraphics[trim=20 21 20 21, clip, width=1\linewidth]{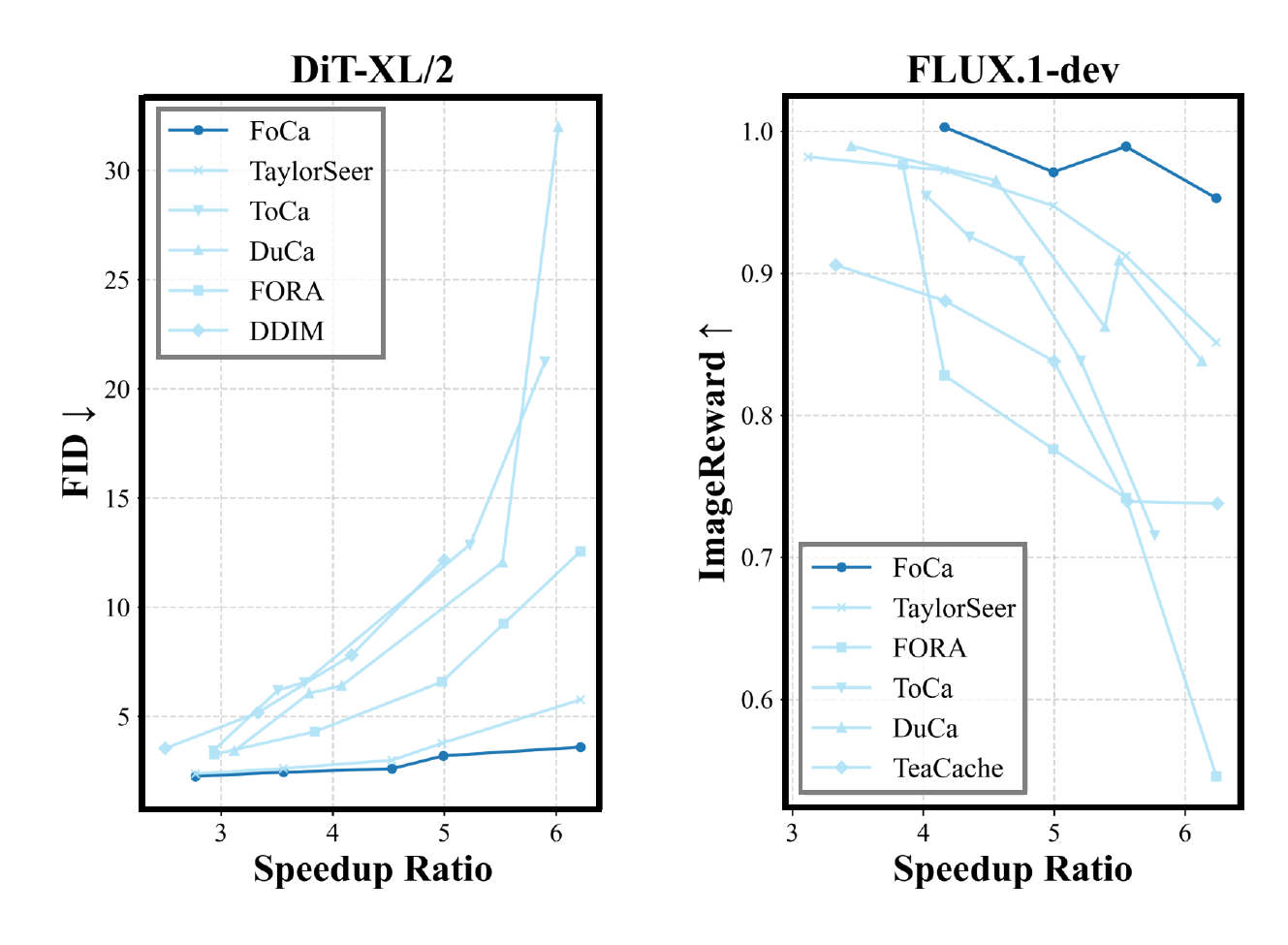}
  \caption{Comparison between previous caching methods and FoCa. FoCa shows significantly better performance.}
  \label{fig:vs1}
\end{figure}

\begin{itemize}
    \item \textbf{ODE Perspective on Feature Caching.}
     We regard the time‑indexed feature tensors as trajectories of a feature-ODE. Framing caching in this perspective lets us apply multistep ODE solvers for prediction, eliminating the need for noisy high-order derivatives while stably propagating hidden states across skipped timesteps.

    \item \textbf{FoCa Framework.} We propose a training-free predictor–corrector framework combining a backward differentiation formula forecasting with Heun calibration, which remains stable even under large acceleration ratios.
     \item \textbf{Outstanding Performance.} We evaluate FoCa across diverse architectures and tasks, including Drawbench on FLUX, Vbench on HunyuanVideo, DiT-XL/2 on ImageNet, and image super-resolution on Inf-DiT using DIV8K. In all settings, FoCa delivers strong acceleration while preserving or enhancing generation quality.
\end{itemize}

\section{Related Works}\label{sec:related_works}

Diffusion models~\citep{sohl2015deep, DDPM} have become a dominant framework for high-fidelity image and video generation. While early implementations primarily adopted U-Net-based architectures~\citep{ronneberger2015unet}, the introduction of Diffusion Transformers (DiTs) has enabled greater scalability and expressiveness. DiTs have since powered numerous state-of-the-art generative systems\citep{chen2023pixartalpha, chen2024pixartsigma, opensora}. Despite progress, the sequential nature of diffusion sampling remains a major obstacle for real-time and large-scale applications. To address this, recent research has focused on two main acceleration strategies: reducing the number of sampling steps and optimizing the efficiency of the denoising network.

\paragraph{Sampling Step Reduction.}
Several approaches aim to shorten the sampling trajectory while preserving output quality. DDIM~\citep{songDDIM} introduces a deterministic sampling formulation that enables fewer steps. The DPM-Solver family~\citep{lu2022dpm, zheng2023dpmsolvervF} applies high-order numerical solvers to more accurately approximate the reverse diffusion process. Rectified Flow~\citep{refitiedflow} constructs direct generative flows via optimal transport, and Consistency Models~\citep{song2023consistency} enforce self-consistency constraints to enable single-step or few-step generation. While effective, these methods generally require redesigning the sampling algorithm or retraining the model, limiting their applicability to pretrained diffusion systems.

\paragraph{Denoising Network Acceleration.}

An alternative line of work focuses on reducing per-step computational cost through model-level optimizations, broadly categorized into model compression and feature caching strategies.

\paragraph{Model Compression.}
Compression-based techniques accelerate inference by simplifying network structures. These include structured pruning~\citep{fang2023structuralpruningdiffusionmodels, zhu2024dipgodiffusionprunerfewstep}, quantization~\citep{kim2025dittoacceleratingdiffusionmodel, li2023qdiffusionquantizingdiffusionmodels}, and token-level reductions such as merging or pruning~\citep{bolya2023tokenmergingfaststable, cheng2025catpruningclusterawaretoken, zhang2025sito}. While these methods can significantly reduce runtime, they typically require retraining or fine-tuning to maintain generation quality, posing challenges for deployment on large-scale pretrained models.

\paragraph{Feature Caching.}
Training-free caching methods have gained attention for their simplicity and compatibility with existing networks. Early approaches such as DeepCache and FasterDiffusion sped up U‑Nets by reusing features across nearby timesteps. Follow‑up work extended caching to transformers: FORA and $\Delta$-DiT store block outputs, while ToCa, DuCa and RAS reuse tokens or regions adaptively.
TaylorSeer replaced direct reuse with Taylor extrapolation, yet it fails to exploit past feature information effectively and can become unstable at larger timesteps because of its higher‑order derivatives. In this work, we propose a more robust forecasting-then-calibrate framework that remains accurate and stable over longer skip intervals, thus delivering higher speed‑ups without retraining or quality loss.

\section{Method}
\subsection{Preliminary}

\paragraph{Diffusion Models.}
Diffusion models operate by simulating a Markovian forward and reverse stochastic process that transitions data between clean images and Gaussian noise. The forward process progressively adds Gaussian noise to the data through a time-dependent variance schedule:
\begin{equation}
    x_t = \sqrt{\alpha_t} x_0 + \sqrt{1 - \alpha_t} \epsilon_t,
\end{equation}
where $\epsilon_t \sim \mathcal{N}(0, I)$ represents isotropic Gaussian noise, and $\alpha_t$ is a monotonically decreasing sequence controlling the noise level. After $T$ timesteps, the sample becomes indistinguishable from pure noise. The reverse process learns to denoise the sample using a neural network $\epsilon_\theta(x_t, t)$ to predict the noise:
\begin{equation}
    x_{t-1} = \frac{1}{\sqrt{\alpha_t}} \left(x_t - \frac{1 - \alpha_t}{\sqrt{1 - \bar{\alpha}_t}} \epsilon_\theta(x_t, t)\right) + \sigma_t \epsilon,
\end{equation}
where $\bar{\alpha}_t = \prod_{s=1}^t \alpha_s$, and $\sigma_t$ controls the noise level during denoising. This reverse process can also be described by a reverse-time stochastic differential equation or its corresponding ordinary differential equation.

\begin{figure*}[t]
  \centering
  \includegraphics[width=\textwidth,
    trim= 20pt 360pt 0pt 10pt,  % 分别是 左 下 右 上
    clip
  ]{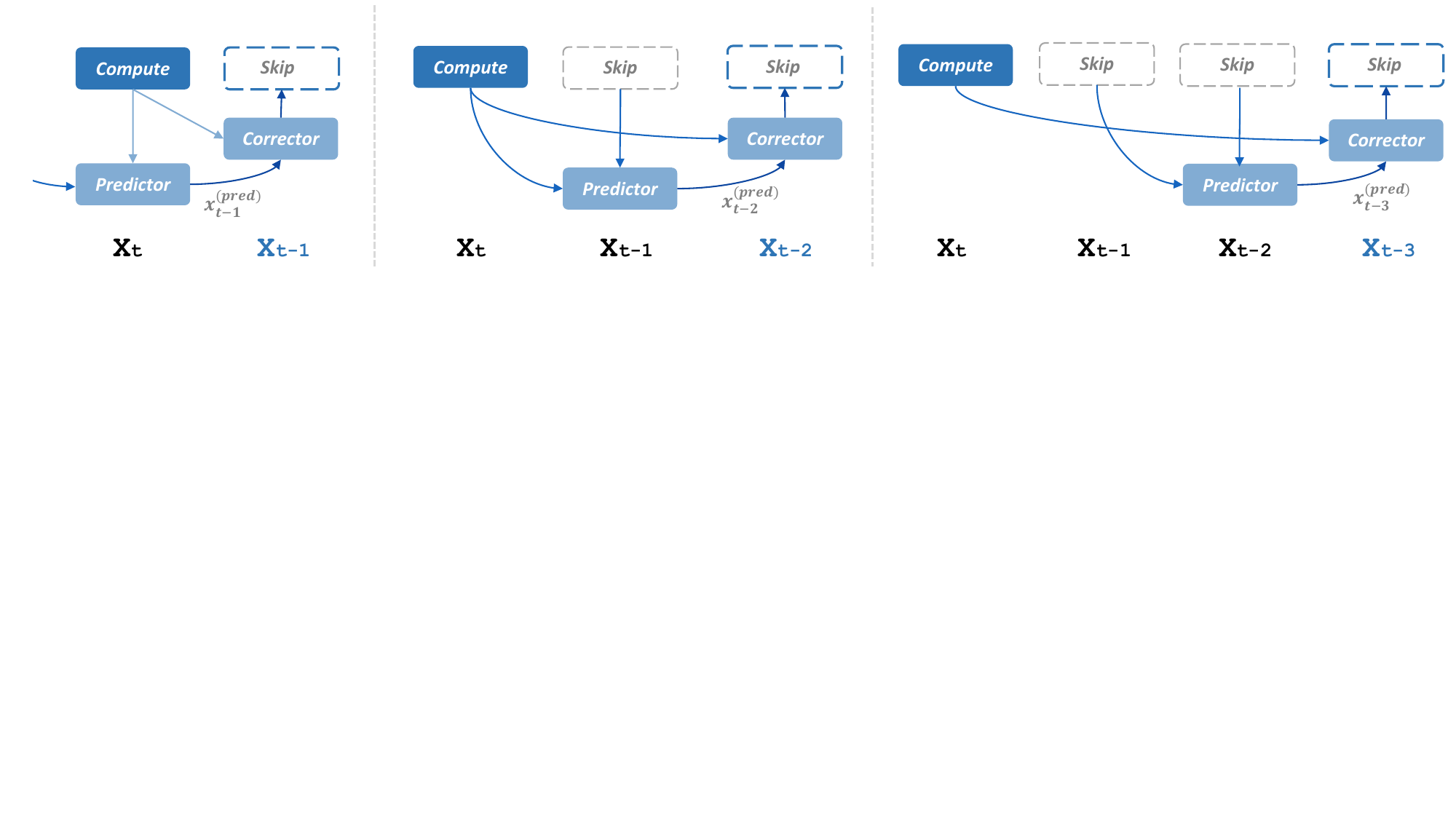}
  \caption{\textbf{The computation of FoCa for $\mathbf{x_{t-1}}$, $\mathbf{x_{t-2}}$, $\mathbf{x_{t-3}} $}, respectively. For each skipping step, FoCa predicts a future hidden state using the two most recent timestep, then applies a Heun corrector that blends the prediction with the most recent fully-computed feature so that enables accurate and stable prediction under large skip intervals. }

  \label{fig:model}
\end{figure*}

\paragraph{Feature Caching in Diffusion Models.}
Feature caching accelerates diffusion inference by avoiding redundant computations across timesteps. Two main strategies exist: \emph{reuse} and \emph{forecast}. The \textbf{reuse-based} method directly substitutes features from a previous timestep:
\begin{equation}
\mathcal{F}(x_{t-k}^l) := \mathcal{F}(x_t^l), \quad \forall k \in [1, N-1],
\end{equation}
achieving $(N{-}1)\times$ acceleration by skipping intermediate steps. However, errors quickly accumulate as the temporal gap increases, degrading output quality. The \textbf{forecast-based} method, exemplified by TaylorSeer, predicts future features via Taylor expansion:
\begin{equation}
\mathcal{F}_{\textrm{pred},m}(x_{t-k}^l) = \mathcal{F}(x_t^l) + \sum_{i=1}^{m} \frac{\Delta^i\mathcal{F}(x_t^l)}{i! \cdot N^i}(-k)^i
\end{equation}
where $\Delta^i \mathcal{F}$ is the $i$-th order finite difference. While more accurate than direct reuse, this method is sensitive to noise under long intervals or complex dynamics, due to the limitations of local smoothness assumptions and the instability of high-order derivatives.

\subsection{Overview of FoCa}

To reduce the high inference cost of diffusion models, we propose \textbf{FoCa}, a predictor-corrector framework designed to accelerate diffusion inference by leveraging historical features for stable and accurate forecasting as shown in Fig.~\ref{fig:model}. Unlike prior methods that rely on local extrapolation and suffer from high-order instability, FoCa integrates multi-step prediction with a lightweight correction mechanism to support longer skip intervals without degrading generation quality. The core workflow consists of three stages:

\begin{itemize}
    \item \textbf{Multi-step Forecasting.} FoCa uses a second-order backward differentiation scheme to predict future hidden features based on the two most recent timesteps. This multi-step strategy captures temporal evolution more accurately than single-step reuse methods and high-order local derivative approximation methods.

    \item \textbf{Heun-style Calibration.} To mitigate potential oscillation or over-extrapolation from the forecasting step, FoCa introduces a correction mechanism that blends the prediction with the latest full-computation feature. This dampens overshoot and stabilizes the inference process under large skipping intervals.

    \item \textbf{Prediction Execution.} Finally, use the corrected feature for the subsequent denoising step unless the next timestep is scheduled for full computation.

\end{itemize}

By combining high-accuracy forecasting with stability-aware correction, FoCa offers a principled and retraining-free solution for accelerating diffusion transformers across image, video, and super-resolution tasks.

\paragraph{An ODE Perspective on Hidden-Feature Evolution.}
During reverse-time denoising in diffusion models, the hidden features evolve smoothly along the sampling trajectory. Let \(x_t^{\,l}=G_l(x_t)\) be the layer-\(l\) activation and \(\mathcal{F}(x_t^{\,l})\) its downstream feature. Since the network blocks are differentiable and \(x_t\) evolves continuously, the composite map \(t \mapsto \mathcal{F}(x_t^{\,l})\) is differentiable; by the chain rule and the probability-flow ODE for \(x_t\),
\begin{equation}
  \frac{d}{dt}\,\mathcal{F}(x_t^{\,l}) \;=\; g_\theta\!\bigl(\mathcal{F}(x_t^{\,l}),\,t\bigr),
  \label{eq:feature_ode_chain}
\end{equation}
where \(g_\theta\) implicitly aggregates Jacobians of preceding layers. Although \(g_\theta\) is intractable, the feature samples \(\{\mathcal{F}(x_{t_k}^{\,l})\}\) on a reverse-time grid permit classical linear multistep integration using only cached values. This insight motivates the use of backward differentiation formula to predict $\mathcal{F}(x_{t_{k+1}}^{\,l})$ from the two most recently stored tensors, thereby skipping full forward passes while retaining accuracy.

\begin{table*}[ht]
    \centering
    \caption{\textbf{Quantitative comparison in text-to-image generation} on FLUX.}
    \setlength\tabcolsep{7.0pt} 
    \small
    
    \begin{tabular}{l | c | c  c | c  c | c | c}
        \toprule
        {\bf Method} & {\bf Efficient} &\multicolumn{4}{c|}{\bf Acceleration} &{\bf Image Reward $\uparrow$} &\bf CLIP$\uparrow$ \\
        \cline{3-6}
        {\bf FLUX.1} & {\bf Attention } & {\bf Latency(s) $\downarrow$} & {\bf Speed $\uparrow$} & {\bf FLOPs(T) $\downarrow$}  & {\bf Speed $\uparrow$} & \bf DrawBench &\bf Score \\
        \midrule
      
        $\textbf{[dev]: 50 steps}$ & {52} & 25.82 & 1.00$\times$ & 3719.50 & 1.00$\times$ & 0.9898 & 32.404 \\ 
        \midrule

        {$60\%$\textbf{ steps}} & {52} & 16.70 & 1.55$\times$ & 2231.70 & 1.67$\times$ & 0.9663 & 32.312 \\

        {$\Delta$-DiT} ($\mathcal{N}=2$) & {52} & 17.80 & 1.45$\times$ & 2480.01 & 1.50$\times$ & 0.9444 & 32.273 \\
        {$\Delta$-DiT} ($\mathcal{N}=3$) & {52} & 13.02 & 1.98$\times$ & 1686.76 & 2.21$\times$ & 0.8721 & 32.102 \\
        \midrule

        {$34\%$\textbf{ steps}} & {52} & 9.07 & 2.85$\times$ & 1264.63 & 3.13$\times$ & 0.9453 & 32.114 \\

        $\textbf{Chipmunk} $ & {52} & 12.72 & 2.02$\times$ & 1505.87 & 2.47$\times$ & 0.9936 & 32.776 \\
        
        $\textbf{FORA}$ $(\mathcal{N}=3)$ & {52} & 10.16 & 2.54$\times$ & 1320.07 & 2.82$\times$ & 0.9776 & 32.266 \\

        $\textbf{\texttt{ToCa}}$ $(\mathcal{N}=6)$ & {56} & 13.16 & 1.96$\times$ & 924.30 & 4.02$\times$ & 0.9802 & 32.083 \\

        $\textbf{\texttt{DuCa}} (\mathcal{N}=5)$ & {52} & 8.18 & 3.15$\times$ & 978.76 & 3.80$\times$ & 0.9955 & 32.241 \\

        $\textbf{TaylorSeer} $ $(\mathcal{N}=4,O=2)$ & {52} & 9.24 & 2.80$\times$ & 1042.27 & 3.57$\times$ & 0.9857 & 32.413 \\

        \rowcolor{gray!20}
        $\textbf{FoCa} $ $(\mathcal{N}=5)$ & {52} & 7.46 & \textbf{3.46}$\times$ & 893.54 & \textbf{4.16$\times$} & \textbf{1.0029} & \bf{32.948} \\
        \midrule

        {$22\%$\textbf{ steps}} & {52} & 6.04 & 4.28$\times$ & 818.29 & 4.55$\times$ & 0.8183 & 31.772 \\

        $\textbf{FORA}$ $(\mathcal{N}=4)$ & {52} & 8.12 & 3.14$\times$ & 967.91 & 3.84$\times$ & 0.9730 & 32.142 \\

        $\textbf{\texttt{ToCa}}$ $(\mathcal{N}=8)$ & {56} & 11.36 & 2.27$\times$ & 784.54 & 4.74$\times$ & 0.9451 & 31.993 \\

        $\textbf{\texttt{DuCa}}$ $(\mathcal{N}=7)$ & {52} & 6.74 & 3.83$\times$ & 760.14 & 4.89$\times$ & 0.9757 & 32.066 \\

        \textbf{TeaCache} $({l}=0.8)$ & {52} & 7.21 & 3.58$\times$ & 892.35 & 4.17$\times$ & 0.8683 & 31.704 \\

        $\textbf{DBcache} $ $(\mathcal{F}=4,B=4)$ & {52} & 6.96 & 3.71$\times$ & 725.40 & 5.12$\times$ & 0.6286 & 31.905 \\

        $\textbf{TaylorSeer} $ $(\mathcal{N}=5,O=2)$ & {52} & 7.46 & 3.46$\times$ & 893.54 & 4.16$\times$ & 0.9768 & 32.467 \\

        \rowcolor{gray!20}
        $\textbf{FoCa} $ $(\mathcal{N}=7)$ & {52} & 6.36 & \textbf{4.05$\times$} & 670.44 & \textbf{5.54$\times$} & \bf{0.9891} & \textbf{32.920} \\
        \midrule

        $\textbf{FORA}$ $(\mathcal{N}=7)$ & {52} & 7.71 & 3.34$\times$ & 670.14 & 5.55$\times$ & 0.7418 & 31.519 \\

        $\textbf{\texttt{ToCa}}$ $(\mathcal{N}=12)$ & {56} & 10.34 & 2.50$\times$ & 644.70 & 5.77$\times$ & 0.7155 & 31.808 \\

        $\textbf{\texttt{DuCa}}$ $(\mathcal{N}=10)$ & {52} & 9.22 & 2.80$\times$ & 606.91 & 6.13$\times$ & 0.8382 & 31.759 \\

        \textbf{TeaCache} $({l}=1.2)$ & {52} & 6.12 & 4.22$\times$ & 669.27 & 5.56$\times$ & 0.7394 & 31.704 \\

        $\textbf{DBcache} $ $(\mathcal{F}=4,B=2)$ & {52} & 6.53 & 3.95$\times$ & 633.44 & 5.87$\times$ & 0.4858 & 31.654 \\
        
        $\textbf{TaylorSeer} $ $(\mathcal{N}=7,O=2)$ & {52} & 6.38 & 4.05$\times$ & 670.44 & 5.54$\times$ & 0.9128 & 32.128 \\

        \rowcolor{gray!20}
        $\textbf{FoCa} $ $(\mathcal{N}=8)$ & {52} & 5.88 & \bf{4.39$\times$} & 596.07 & \bf{6.24$\times$} & \textbf{0.9502} & \textbf{32.706} \\

        \bottomrule
    \end{tabular}
    \label{table:FLUX-Metrics}
\end{table*}

\paragraph{BDF2 Forecasting.}
The second-order backward differentiation formula (BDF2) is a linear multistep method. It approximates the derivative at the next step, $k+1$, using the latest two points ($\mathcal{F}(x_{k-1}^l), \mathcal{F}(x_k^l)$):
\begin{equation}
\mathcal{F}^{(1)}(x_{k+1}^l) = \frac{3\mathcal{F}(x_{k+1}^l) - 4\mathcal{F}(x_k^l) + \mathcal{F}(x_{k-1}^l)}{2h_{k}}.
\end{equation}
This formula is implicit because the right-hand side depends on the unknown $\mathcal{F}(x_{k+1}^l)$ through the function $\mathcal{F}^{(1)}$. To create a practical, explicit forecasting scheme, we approximate the required future derivative $\mathcal{F}^{(1)}(x_{k+1}^l)$ by extrapolating from the most recent step's derivative, $\mathcal{F}^{(1)}(x_{k}^l)$. This yields our predictor:

\begin{equation} \label{eq:bdf2_explicit}
\hat{\mathcal{F}}(x_{k+1}^l) = \frac{4}{3}\mathcal{F}(x_k^l) - \frac{1}{3}\mathcal{F}(x_{k-1}^l) + \frac{2h_{k}}{3} \mathcal{F}^{(1)}(x_{k}^l).
\end{equation}

\paragraph{Heun Calibration.}
We enhance our BDF2 forecast by applying a Heun-style corrector step. The Heun method, also known as the explicit trapezoidal rule, improves the prediction by averaging the slopes at the beginning and the predicted end of the interval. First, we compute the BDF2 forecast $\hat{\mathcal{F}}(x_{k+1}^l)$ using Eq. \ref{eq:bdf2_explicit}. Then, we apply the Heun corrector to obtain the final, more accurate state $\mathcal{F}_{c}(x_{k+1}^l)$:
\begin{equation}
\mathcal{F}_{c}(x_{k+1}^l) = \mathcal{F}(x_k^l) + \frac{h_{k}}{2} [ \mathcal{F}^{(1)}(x_{k-N}^l) + \mathcal{F}^{(1)}(x_{k+1}^l)].
\end{equation}
Where $\mathcal(x_{k-N}^l)$ is the latest fully computed step.

\noindent\textbf{Proposition 1 (FoCa ensures stable prediction under large interval, proof in Appendix A.1).}
\textit{
Given a smooth feature evolution $\mathcal{F}(x_t^l)$ in diffusion models, FoCa achieves a prediction error bound that is independent of the number of skipped steps $k$:
\begin{equation}
 \| \hat{\mathcal{F}}(x_k^l) - \mathcal{F}^*(x_k^l) \| \leq \left( \frac{1-\rho^k}{1-\rho} \right) \tau_{\max} \leq \frac{\tau_{\max}}{1-\rho} 
\end{equation}
In contrast, the error of TaylorSeer and reuse methods increases with $k$, leading to degraded accuracy in long-range prediction. This highlights FoCa's robustness under aggressive caching and large acceleration ratios.
}

\paragraph{Practical Impact.}
This predictor-corrector framework combines the strengths of both methods. The BDF2 predictor leverages historical information for initial prediction. The Heun corrector then refines this prediction by incorporating the most recent fully computed step, effectively damping oscillations and reducing error accumulation. This hybrid approach allows larger acceleration ratios while maintaining high-fidelity generation, outperforming methods that rely on either single-step reuse or less accurate high-order derivative extrapolation schemes.

\section{Experiments}

\begin{table*}[htbp]
\centering
\caption{\textbf{Quantitative comparison in text-to-video generation} on VBench.}

\setlength\tabcolsep{12pt} 
\small

\begin{tabular}{l | c | c  c | c  c | c }
    \toprule
    {\bf Method} & {\bf Efficient} & \multicolumn{4}{c|}{\bf Acceleration} & {\bf VBench $\uparrow$} \\
    \cline{3-6}
    {} & {\bf Attention} & {\bf Latency(s) $\downarrow$} & {\bf Speed $\uparrow$} & {\bf FLOPs(T) $\downarrow$} & {\bf Speed $\uparrow$} & \bf Score(\%) \\
    \midrule
  
    $\textbf{Original}$ 
                           & {\ding{52}}  & {334.96} & {1.00$\times$} & {29773.0} & {1.00$\times$} & {80.66} \\ 
    \midrule

    $\textbf{DDIM-22\%}$  
                           & {\ding{52}}  & {87.01} & {3.85$\times$} & {6550.1} & {4.55$\times$} & {78.74} \\
    $\textbf{FORA}$ $(\mathcal{N}=5)$ 
                           & {\ding{52}}  & {83.78} & {4.00$\times$} & {5960.4} & {5.00$\times$} & {78.83} \\

    $\textbf{\texttt{ToCa}}$ $(\mathcal{N}=5)$ 
                           & {\ding{56}}  & {93.80} & {3.57$\times$} & {7006.2} & {4.25$\times$} & {78.86} \\

    $\textbf{\texttt{DuCa}}$ $(\mathcal{N}=5)$ 
                           & {\ding{52}}  & {87.48} & {3.83$\times$} & {6483.2} & {4.62$\times$} & {78.72} \\
    \textbf{TeaCache}  $({l}=0.4)$       & {\ding{52}}  & 70.43 & 4.76$\times$ & 6550.1 & 4.55$\times$ & 79.36 \\
    
    \textbf{TeaCache}   $({l}=0.5)$    & {\ding{52}} & 61.47 & 5.45$\times$ & 5359.1 & 5.56$\times$ & 78.32 \\
    
    $\textbf{TaylorSeer}$ $(\mathcal{N}=7,O=1)$ 
                           & {\ding{52}}  & {72.76} & {4.60$\times$} & {4611.3} & {6.45$\times$} & {79.37} \\

    \rowcolor{gray!20}
    $\textbf{FoCa}$ $(\mathcal{N}=7)$ 
                           & {\ding{52}}  & {72.18} & {4.64$\times$} & \bf{4611.3} & {\textbf{6.45}$\times$} & \textbf{79.68} \\

    \bottomrule
\end{tabular}
\label{table:HunyuanVideo-Metrics}
\end{table*}

\noindent\textbf{Experiment Settings.}  
We evaluate our method on four representative diffusion-based generative models: the text-to-image model \textbf{FLUX.1-dev}~\cite{flux2024}, the super-resolution model \textbf{Inf-DiT}~\cite{infdit2024}, the text-to-video model \textbf{HunyuanVideo}~\cite{sun_hunyuan-large_2024}, and the class-conditional image generation model \textbf{DiT-XL/2}~\cite{DiT}. For text-to-image generation, we follow the DrawBench~\cite{sahariaPhotorealisticTexttoImageDiffusion2022} protocol and evaluate photorealism and text alignment using ImageReward~\cite{xuImageRewardLearningEvaluating2023} and CLIP Score~\cite{CLIPScore}.  
For text-to-video generation, we adopt the VBench~\cite{VBench} benchmark, which assesses video quality across multiple human-aligned dimensions.  
For class-conditional image generation, we evaluate on ImageNet~\cite{Imagenet} using FID-50k~\cite{FID50K} and Inception Score (IS) as standard quality metrics.  
For super-resolution, we use DIV8K~\cite{div8k} and measures the quality with PSNR and SSIM. Please refer to Appendix for details.

\begin{figure}[htbp]
  \centering
  \includegraphics[trim=10 0 10 0, clip,width=1\linewidth]{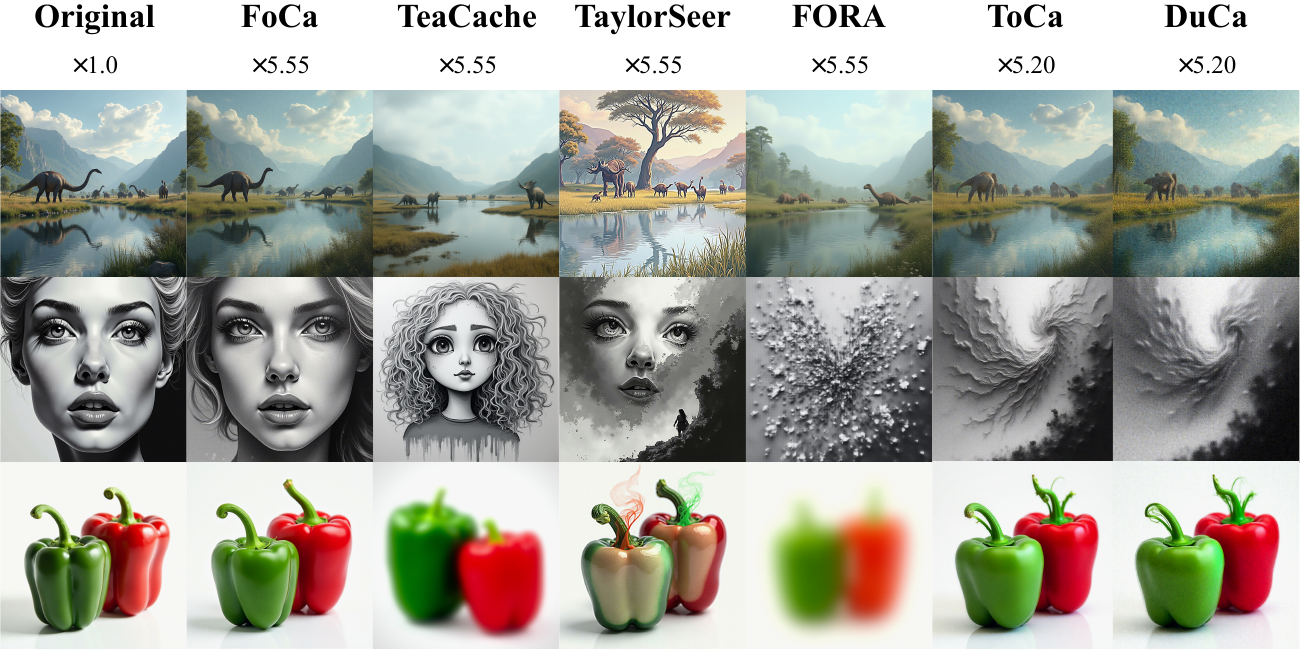}
  \caption{Visual comparison of 5.5$\times$ accelerated FLUX.}
  \label{fig:flux}
\end{figure}

\noindent\textbf{Results on Text-to-Image Generation.}
As summarized in Table~\ref{table:FLUX-Metrics}, our method achieves a consistently superior speed–quality trade-off on FLUX.1-dev across all acceleration levels.  With a moderate cache interval ($\mathcal{N}=5$), FoCa attains the highest ImageReward score of \textbf{1.0029} while achieving a 4.16$\times$ FLOPs reduction, outperforming all reuse-based baselines, and the prediction-based method TaylorSeer (0.9857 at 3.57$\times$). Under higher compression ($\mathcal{N}=7$), FoCa retains a strong ImageReward of \textbf{0.9891} at a 5.54$\times$ speedup, surpassing TaylorSeer's performance at $\mathcal{N}=5$, $O=2$ (0.9857), while other baselines degrade notably. At the most aggressive setting ($\mathcal{N}=8$), FoCa still achieves \textbf{0.9502} ImageReward at 6.24$\times$ acceleration, significantly outperforming other methods. Qualitative comparisons in Figure~\ref{fig:flux} further illustrate that FoCa preserves better image quality, even under large skipping intervals.

\begin{figure*}[t]
  \centering
  \includegraphics[trim=15 10 15 10, clip, width=\textwidth]{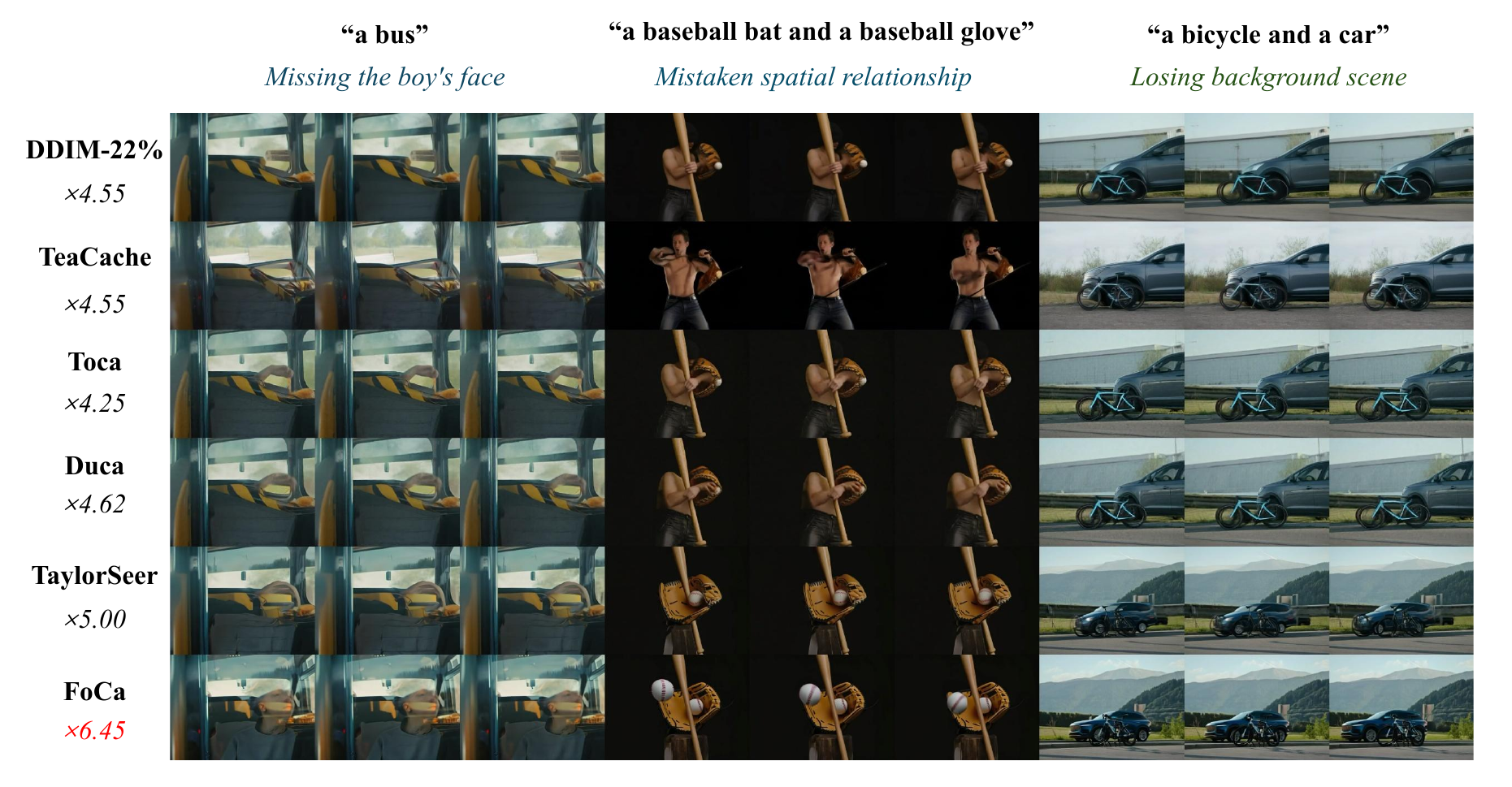}
  \caption{Visualization of different caching methods on HunyuanVideo. FoCa maintains high-quality generation under the higher acceleration ratio, while other methods suffer from issues such as missing details, mistaken spatial relationships, and losing the background scene.}
  \label{fig:video}
\end{figure*}

\noindent\textbf{Results on Text-to-Video Generation.}
Table~\ref{table:HunyuanVideo-Metrics} demonstrates that FoCa delivers the best overall trade-off on HunyuanVideo. With a cache window of $\mathcal{N}=7$, FoCa holds a \textbf{6.45}$\times$\ speedup ratio, while achieving the highest VBench score of \textbf{79.68}\%. In particular, compared to the original sampler (80.66\%), FoCa attains near-lossless quality with more computational savings $6.45\times$. Qualitative comparisons on Fig.~\ref{fig:video} further confirm that FoCa preserves video quality, spatial relationship, background scenery better.

\begin{table}[ht]
    \centering
    \caption{\textbf{Quantitative comparison in class-to-image generation} on ImageNet with \text{DiT-XL/2}.}
    \setlength\tabcolsep{4pt}
    \tiny
    
    \begin{tabular}{l | c c c | c c}
    \toprule
    \bf Method
      & \bf Latency(s) $\downarrow$
      & \bf FLOPs(T) $\downarrow$
      & \bf Speed $\uparrow$
      & \bf FID $\downarrow$
      & \bf IS $\uparrow$ \\
    \midrule
    \textbf{DDIM-50 steps}
      & 0.428 & 23.74 & 1.00$\times$ & 2.32  & 241.25 \\
    \midrule
    \textbf{DDIM-25 steps}
      & 0.230 & 11.87 & 2.00$\times$ & 3.18  & 232.01 \\
    \textbf{L2C} ($NFE=30$)
      & 0.281 & 11.55 & \textbf{2.05}$\times$ & 2.61  & 237.83 \\
    $\Delta$-DiT ($\mathcal{N}=2$)
      & 0.246 & 18.04 & 1.31$\times$ & 2.69  & 225.99 \\
    $\Delta$-DiT ($\mathcal{N}=3$)
      & 0.173 & 16.14 & 1.47$\times$ & 3.75  & 207.57 \\
    \rowcolor{gray!20}
    \textbf{FoCa} ($\mathcal{N}=2$)
      & 0.238 & 12.35 & 1.92$\times$ & \textbf{2.17}  & \textbf{239.94} \\
    \midrule
    \textbf{DDIM-20 steps}
      & 0.191 &  9.49 & 2.50$\times$ & 3.81  & 221.43 \\
    \textbf{FORA} ($\mathcal{N}=3$)
      & 0.197 &  8.58 & 2.77$\times$ & 3.55  & 229.02 \\
    \texttt{ToCa} ($\mathcal{N}=3$)
      & 0.216 & 10.23 & 2.32$\times$ & 2.87  & 235.21 \\
    \texttt{DuCa} ($\mathcal{N}=3$)
      & 0.208 &  9.58 & 2.48$\times$ & 2.88  & 233.37 \\
    \textbf{SmoothCache} ($\alpha=0.22$)
      & 0.251 &  8.57 & 2.77$\times$ & 4.15  & 231.71 \\
    \textbf{TaylorSeer} ($\mathcal{N}=3$)
      & 0.248 &  8.56 & 2.77$\times$ & 2.35  & 236.22 \\
    \rowcolor{gray!20}
    \textbf{FoCa} ($\mathcal{N}=3$)
      & 0.250 &  8.56 & \textbf{2.77}$\times$ & \textbf{2.25}  & \textbf{237.70} \\
    \midrule
    \textbf{DDIM-12 steps}
      & 0.128 &  5.70 & 4.17$\times$ & 7.80  & 184.50 \\
    \textbf{FORA} ($\mathcal{N}=5$)
      & 0.149 &  5.24 & 4.53$\times$ & 6.58  & 193.01 \\
    \texttt{ToCa} ($\mathcal{N}=6$)
      & 0.163 &  6.34 & 3.75$\times$ & 6.55  & 189.53 \\
    \texttt{DuCa} ($\mathcal{N}=5$)
      & 0.154 &  6.27 & 3.78$\times$ & 6.06  & 198.46 \\
    \textbf{TaylorSeer} ($\mathcal{N}=5$)
      & 0.186 &  5.24 & 4.53$\times$ & 2.74  & \textbf{231.42} \\
    \rowcolor{gray!20}
    \textbf{FoCa} ($\mathcal{N}=5$)
      & 0.183 &  5.24 & \textbf{4.53}$\times$ & \textbf{2.60}  & 230.76 \\
    \bottomrule
    \end{tabular}
    \label{table:DiT_Metrics}
\end{table}

\noindent\textbf{Results on Class-to-Image Generation.}
On DiT-XL/2, FoCa establishes a stronger quality–efficiency frontier than other training-free baselines, and it also surpasses the training-based L2C at comparable budgets. With interval ($\mathcal{N}=2$), FoCa' FID even improves over the original 50-step DDIM at 1.92$\times$ speedup; at $\mathcal{N}=3$ it achieves FID 2.25 (2.77$\times$), outperforming TaylorSeer FID 2.35. Under higher compression ($\mathcal{N}=5$), FoCa maintains FID 2.60 at 4.53$\times$, whereas most method collapse.

\noindent\textbf{Results on Image-Super-Resolution.}
We have evaluated our method and the other caching methods for image super-resolution. On Inf-DiT, FoCa yields a \textbf{3.17$\times$} speedup with even better PSNR of \textbf{31.03\,dB}. In contrast, TaylorSeer achieves only 29.66\,dB, confirming that noisy finite-difference forecasting degrades high-frequency detail. These results also demonstrate that FoCa's robustness beyond text-conditional synthesis.

\begin{table}[ht]
    \centering
    \caption{\textbf{Quantitative comparison in on Inf-DiT} for genetative image super-resolution from 512$\times$512 to 2048$\times$2048.}
    \setlength\tabcolsep{4.5pt}
    \tiny
    
    \begin{tabular}{l | c c c | c c}
    \toprule
    \bf Method
      & \bf Latency(s) $\downarrow$
      & \bf FLOPs(T) $\downarrow$
      & \bf Speed $\uparrow$
      & \bf PSNR $\uparrow$
      & \bf SSIM $\uparrow$ \\
    \midrule
    \textbf{Original Model}
      & 451.3 & 124004 & 1.00$\times$ & 30.85 & \textbf{0.829} \\
    \textbf{AB-Cache}
      & 197.2 & 45177 & 2.74$\times$ & 29.77 & 0.801 \\
    \textbf{TaylorSeer} ($\mathcal{N}=4$)
      & 199.5 & 45184 & 2.74$\times$ & 29.67 & 0.805 \\
    \rowcolor{gray!20}
    \textbf{FoCa} ($\mathcal{N}=5$)
      & 161.7 & 39096 & \textbf{3.17}$\times$ & \textbf{31.03} & {0.814} \\
    \bottomrule
    \end{tabular}
    \label{table:DiT_Metrics}
\end{table}

\begin{figure}[H]
  \centering
  \includegraphics[trim=25 20 25 25, clip, width=1\linewidth]{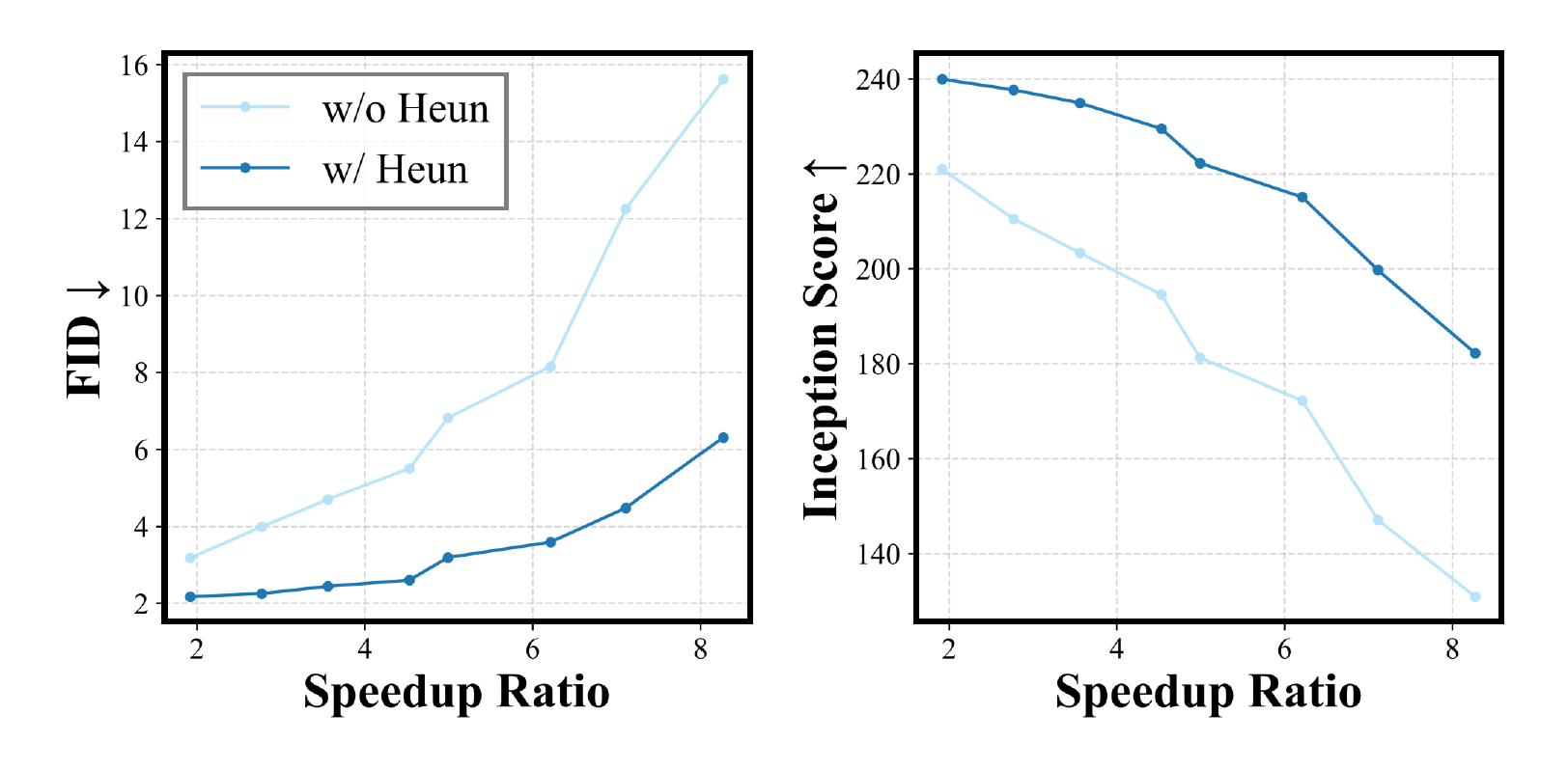}
  \caption{Ablation study of Heun Correction. Adding the Heun step consistently reduces FID and boosts Inception Score across all speedup ratios.}
  \label{fig:ablation}
\end{figure}

\noindent\textbf{Ablation Study.}
We evaluate the impact of Heun correction in FoCa on DiT-XL/2. As shown in Figure~\ref{fig:ablation}, we compare the BDF2-only with FoCa at varying activation intervals. While BDF2-only forcasting deteriorates rapidly under large skip intervals, FoCa consistently achieves lower FID and higher Inception Score, demonstrating improved stability and accuracy. These results highlight the importance of integrating stability-aware correction into multi-step caching schemes, particularly on aggressive acceleration.

\begin{figure}[t]
  \centering
  \includegraphics[trim=25 10 0 20, clip, width=1\linewidth]{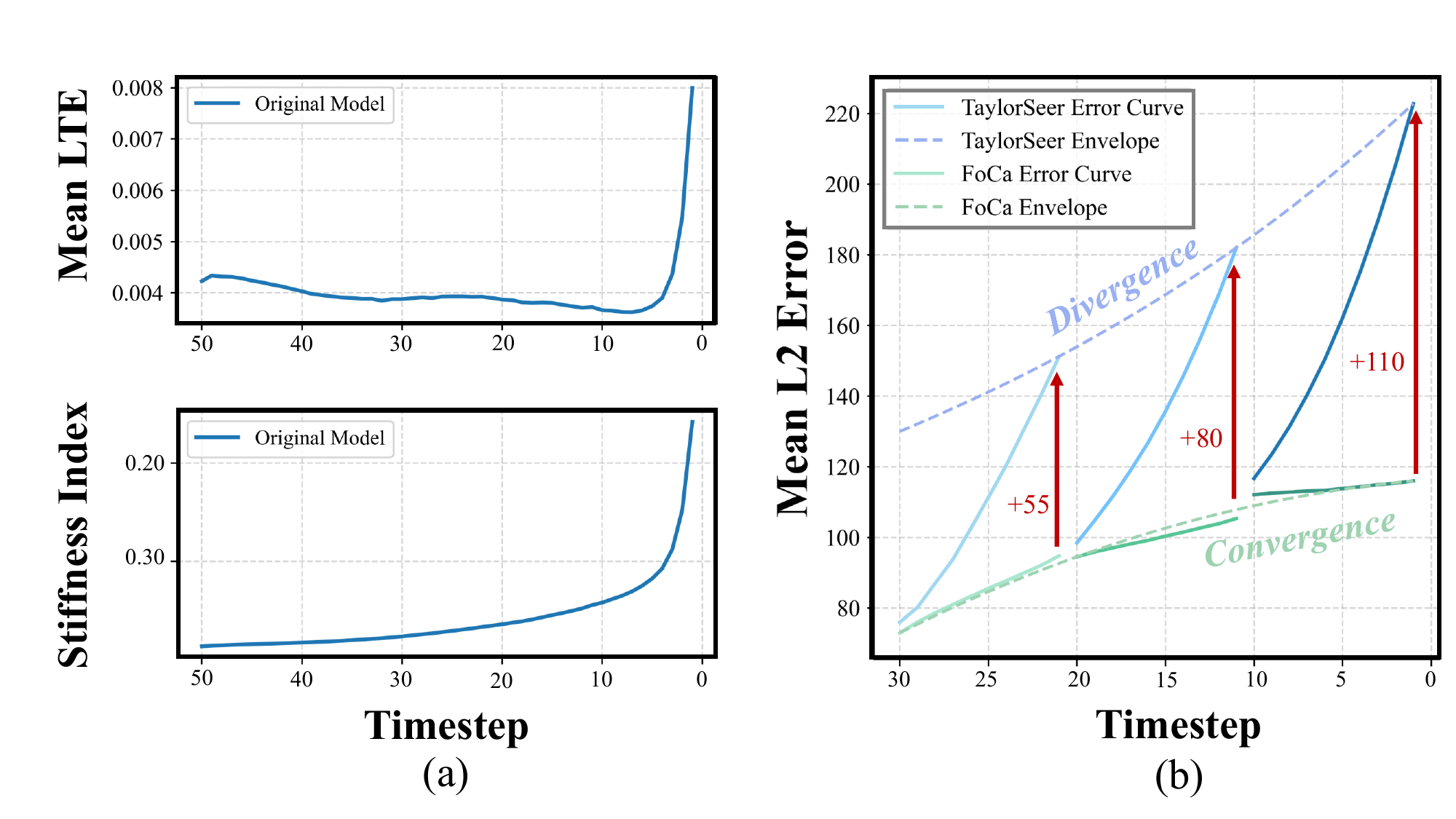}
  \caption{\textbf{Stiffness Analysis}: (a) LTE and Stiffness Index of FLUX across timesteps, revealing a sharp rise in the late stage that indicates a numerically stiff region. LTE denotes local truncation error, which measures numerical integration accuracy. The Stiffness Index measures the numerical stiffness of the model. (b) Comparison of ten-step forecasting error starting from various timesteps. }
  \label{fig:stiffness}
\end{figure}

\noindent\textbf{Analysis on the stiffness of feature trajectories.}
In numerical analysis, a differential equation is considered stiff when certain components evolve at vastly different timescales, causing instability for standard integration methods unless extremely small step sizes are used. In this work, we model the evolution of hidden features in diffusion models as an ODE system and solve it using a BDF2-Heun predictor-corrector scheme. This method is particularly well-suited for stiff ODEs, making it a natural fit for handling the challenging dynamics of diffusion sampling. To explore the applicability of our method, we investigate the stiffness of the hidden feature trajectories. Our stiffness diagnosis in Fig.~\ref{fig:stiffness}(a) reveals a late-stage region (from step 10 to 0) where LTE and Stiffness Index both rise sharply. This indicates high sensitivity to step-size perturbations and numerical instability, which we define as the stiff zone in the evolution of hidden states. To further analyze, Fig.~\ref{fig:stiffness}(b) shows forecasting error under controlled prediction horizons. Specifically, we execute full computation up to timestep 30, 20, and 10, and forecast the next 10 steps using either TaylorSeer or FoCa. As prediction begins deeper into the stiff zone, TaylorSeer exhibits rapidly diverging errors due to over-extrapolation, while FoCa maintains bounded error growth through stable multi-step aggregation and Heun-style correction. This results in convergent error curves and consistent generation quality even under aggressive skipping.

\section{Conclusion}
We presented \textbf{FoCa}, a training-free acceleration framework that reinterprets hidden feature state prediction as solving an ODE and couples BDF2 predictor with a Heun calibrator. This predictor–corrector design utilizes historical features effectively and expands the stability region into the stiff zone where existing caching methods collapse. Performing \emph{near-lossless} speedups in diverse settings: \textbf{ 5.50$\times$} on FLUX, \textbf{6.45$\times$} on HunyuanVideo, \textbf{3.17$\times$} on Inf-DiT and state-of-the-art trade-offs on ImageNet with DiT-XL/2 (up to 4.53$\times$ with superior FID). FoCa provides a novel acceleration perspective from the view of ODE in the feature space. %As a plug-and-play cache strategy, FoCa requires no retraining or architecture changes and can be combined with orthogonal accelerators such as pruning, quantization. Future work includes adaptive step-size control within the predictor–corrector loop, extending FoCa to other modalities (audio, 3D, multimodal agents), and theoretically tightening global error bounds under realistic network noise. We believe FoCa offers a principled, stable, and practical path toward deploying diffusion transformers under strict latency and compute budgets.

\bibliography{aaai2026}

\end{document}